# A Fuzzy View on k-Means Based Signal Quantization with Application in Iris Segmentation

Nicolaie Popescu-Bodorin, *IEEE Member,*
http://fmi.spiruharet.ro/bodorin/

*Abstract* — **This paper shows that the k-means quantization of a signal can be interpreted both as a crisp indicator function and as a fuzzy membership assignment describing fuzzy clusters and fuzzy boundaries. Combined crisp and fuzzy indicator functions are defined here as natural generalizations of the ordinary crisp and fuzzy indicator functions, respectively. An application to iris segmentation is presented together with a demo program.**

## I. Introduction

RELATIVELY few iris segmentation techniques have been reported in the last two decades. In the classical iris segmentation procedures, like those in Wildes's [1] and Daugman's approaches [2]-[10], iris segmentation means fitting (nearly) circular contours by solving 3-dimensional optimization problems to find *a radius* and *two center coordinates* via gradient ascent or by using edge detectors and Hough transform [11] or by iterating active contours [9,10]. For more details, we would like to refer to Bowyer et al. for a survey of iris recognition [12].

All of the previous iris segmentation approaches are 3-dimensional optimization problems. They also assume that the segmentation is done before iris unwrapping. In this context two questions must be answered:

*Is it possible to formulate the finding of the limbic boundary as a 1-dimensional optimization problem or as a search in a 1-dimensional parameter space? If yes, would the resulting iris segments be accurate enough to guarantee strong recognition results?*

We affirmatively answer these questions by giving, in the same time, four reasons to work with circular approximation of the iris:
i) It is clear that only by knowing the center coordinates and by unwrapping the iris region in the first place, limbic boundary finding could become a 1-dimensional search for a radius i.e. a search for a line within the unwrapped iris region. Consequently, assuming a rough approximation of the actual iris as a circular ring concentric with the pupil is a choice [13]-[15] which guarantees an affirmative answer to the first question from above.
ii) An anatomic argument for using circular approximation of the iris is that since the pupillary boundary is nearly circular, there must be a circular concentric iris ring controlling the pupil movements. Such a circular iris ring is expected to play the most important role in iris recognition, despite the fact that it appears to be a rough approximation of the actual iris.
iii) A system requirement sustaining the use of concentric circular iris ring is that the segmentation routine must be fast and energy-efficient. Nearly lossless unwrapping of the iris can be computed using a polar or a bipolar coordinate transform, depending on the type assumed for the iris: concentric or eccentric circular ring. The latter is computationally more expensive than the former because the eccentricity varies from a sample to another and consequently, one bipolar mapping must be (re)computed for each sample (eye image). When the iris ring is assumed to be concentric, the polar mapping is computed only once for all samples, during program initialization.
iv) At last, but not the least, a practical argument for using circular approximation of the iris is given by the quality of the recognition results presented in [13], [14] and by the iris segmentation results illustrated in [15].

A harder question regarding the Circular Fuzzy Iris Segmentation procedure [13]-[15] is the following: *why the various operations (within the segmentation procedure) are needed or expected to work well?*

The short answer to the above question came as a result of our experimental works and is stated here as a principle: *detecting a certain feature of a signal (of an image) is always a matter of finding a suitable quantization space and a suitable quantization function to enhance the target feature against 'unwanted noise' (against the surrounding neighbours in the feature space). In the best case scenario feature discovery would be nothing more than a well chosen binary encoding (compression) of the feature space.*

Another four principles of k-means optimal signal

Nicolaie Popescu-Bodorin is with the Mathematics and Computer Science Department - 'Spiru Haret' University of Bucharest (România), where he teaches Computational Logic and Artificial Intelligence Labs. As a PhD Candidate at Mathematics and Computer Science Department - University of Pitesti (România), he works in the field of Digital Signal/Image Processing and specifically on Iris Recognition related subjects. Correspondence address: Nicolaie Popescu-Bodorin, OP 19, CP 77, Bucharest 3, RO.

quantization can be found in [16][1] together with the Fast k-Means Image Quantization algorithm (FKMQ).

A longer and more detailed answer to the last question from above will be given further in this paper.

## II. COMBINED CRISP AND FUZZY INDICATORS OF A DISJOINT REUNION

Generally speaking, a segmentation technique working on discrete signals is a semantic operator encoding the input signal using a finite set of labels (symbols) which are somehow meaningful in human understanding of the input signal. The first difficulty in interpreting a segmentation as being fuzzy is the lack of the instruments that could enable us to view the result of a segmentation as a crisp or as a fuzzy membership function defined from the input signal to a collection of segments encoded as a list of *arbitrary symbols*, possibly *non-numeric*, and more often found *outside* [0, 1] interval. This section is meant to work around this issue by extending the definition of the ordinary crisp and fuzzy indicator functions to cover the above described situation.

In fuzzy set theory [18], a membership function that only takes binary values is called a crisp indicator function. We extend the meaning of this definition by making the following considerations: a crisp indicator is, in fact, the ordinary indicator function of an ordinary sub-set within a set:

$$I_A : X \to \{0,1\}; \forall a \in X, I_A(a) = logical(a \in A) \quad (1)$$

For any sub-set A of X, $X = A \cup \overline{A}$ (where $\overline{A}$ denotes the complement of A in X), hence we may consider that the crisp indicator of A is nothing more than an encoding (in two symbols) of a disjoint cover of X containing two sets: A and $\overline{A}$ (regardless the nature or the values of those two symbols and the nature of the sets A and X). Consequently, it is naturally to define *combined crisp indicator of a disjoint reunion*:

$$X = \bigcup_{j=1}^{n} A_j, \quad (2)$$

as being the sum:

$$CCI_X = \sum_{j=1}^{n} j * I_{A_j}, \quad (3)$$

or more generally, as follows:

$$\forall k = \overline{1,n}, \forall a \in A_k, CCI_X(a) = S\left(\sum_{j=1}^{n} j * I_{A_j}(a)\right) = s_k, \quad (4)$$

where $S = \{s_k\}_{k=\overline{1,n}}$ is a sequence of distinct symbols. It means that a combined crisp indicator of a disjoint reunion is unique up to a bijective correspondence between the sequences of symbols that are used to encode the memberships to each set within the reunion. Hence, if X is restricted to R, the combined crisp indicator of a disjoint cover of X is exactly the equivalence class of all step functions that can be defined using the sets of that cover. If X is a discrete signal, then we talk in terms of discrete step functions. Consequently, any discrete step function (and in particular, any k-means quantization of a discrete signal) is equivalent (in the above defined sense) to a combined crisp indicator (3). Therefore, it doesn't really matter what symbols (or values) are used to encode the crisp indicator function. Chromatic k-means centroids and cluster indices {1,…,n} are both equally suitable to encode a crisp indicator function describing the k-means clusters.

---

[1] In [16], some considerations regarding the speed of FKMQ are already outdated by the newer and faster implementation [17]. Also, the iris segmentation procedure proposed in [16] was temporarily abandoned for the following reason: despite its accuracy in finding the iris segment available in the eye image (see Fig.6-7 in [16]), 'guessing' the best eccentric circular ring that matches the available iris segment proved to be a tough challenge (an ill-posed inverse problem with 6 variables) and therefore, was impossible to formulate recognition results based on that segmentation technique. Still, future solutions to this problem are not excluded. On the other hand, it must be mentioned that the difference between the segmentation procedure proposed in [16] and Circular Fuzzy Iris Segmentation [13]-[15] is that the latter searches directly for the line approximating limbic boundary in the unwrapped iris region (it searches for a line number in a different and smaller feature space).

The ordinary crisp indicator of a set is unique (up to a bijection, as described above), but the ordinary fuzzy membership assignment functions are not. The *combined fuzzy indicators of a disjoint reunion* inherit this property and they are defined here as follows: given a *combined crisp indicator* of the form (3), any monotone function $CFI_X$ satisfying the relation:

$$[CFI_X] = CCI_X, \quad (5)$$

where [·] denotes the integer part function, is a *combined fuzzy indicator* of the given disjoint reunion (2). In other words, the function:

$$FIB_X = 2 * abs(CFI_X - CCI_X) \quad (6)$$

is an *ordinary fuzzy indicator of the boundaries* between the sets of the reunion (2). Naturally, the *combined crisp and fuzzy indicators* (3, 5) of a disjoint reunion (2) and the *ordinary fuzzy indicator of the boundaries* (6) form an interdependent triplet.

## III. CIRCULAR FUZZY IRIS RING AND CIRCULAR FUZZY LIMBIC BOUNDARY

Finding the pupil [11],[13],[14] enables us to unwrap a circular pupil-concentric region of the eye image (Fig.1.a) in polar coordinates (Fig.1.b), to localize the limbic boundary in the rectangular unwrapped eye image (Fig.1.c), and to obtain an iris segment as in Fig.1.e.

**Circular Fuzzy Iris Segmentation (CFIS, N. Popescu-Bodorin):**
INPUT: the eye image IM;
 1. Apply RLE-FKMQ Based Pupil Finder procedure;
 2. Unwrap the eye image in polar coordinates (UI - Fig.1.b);
 3. Stretch the unwrapped eye image UI to a rectangle (RUI - Fig.1.c);
 4. Compute three column vectors: A, B, C, where A and B contain the means of the lines within UI and RUI, respectively. C is the mean of the lines within the [A B] matrix;
 5. Compute P, Q, R as being 3-means quantizations of A, B, C;
 6. For each line of the unwrapped eye image count the votes given by P, Q and R. All lines voted at least twice as members of an iris band are assumed to belong to the actual iris segment. Find limbic boundary and extract iris segment (Fig.1.d, Fig.1.e);
OUTPUT: pupil center/radius, line number of the circular fuzzy limbic boundary, circular fuzzy iris segment;
END.

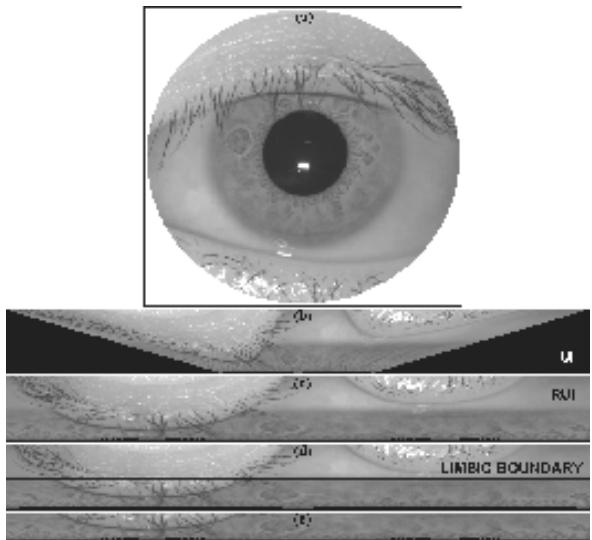

Fig.1. Iris segmentation stages (CFIS)

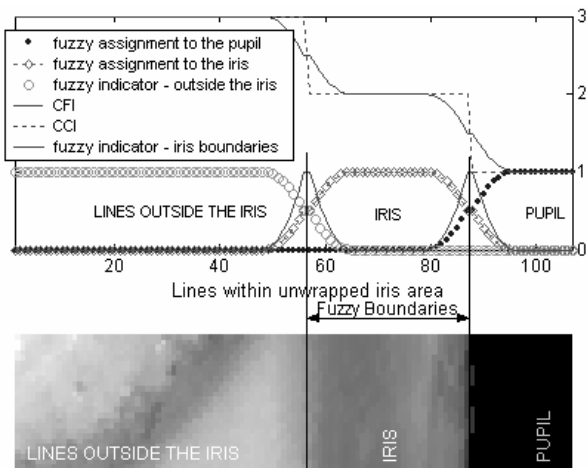

Fig.2. Fuzzy iris segment and fuzzy iris boundaries

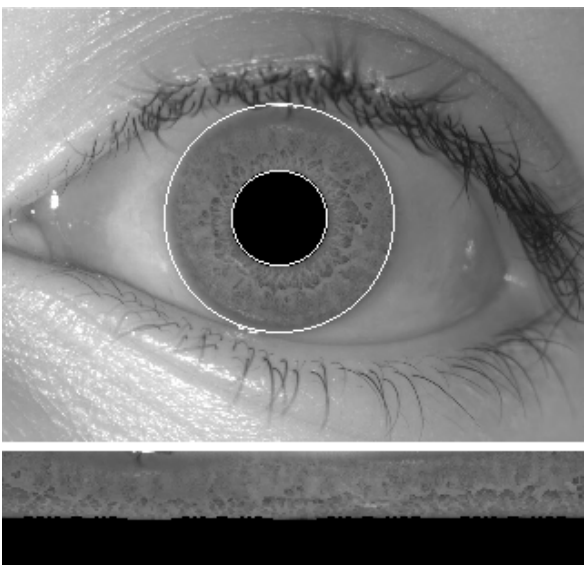

Fig.3. Circular Fuzzy Iris Segmentation Demo Program, http://fmi.spiruharet.ro/bodorin/arch/cffis.zip

Fig.1 shows iris segmentation stages. The transform from Fig1.a to Fig.1.b is a lossless pixel-to-pixel transcoding. The unwrapped iris region is further stretched and interpolated in order to obtain rectangular unwrapped iris (RUI – Fig.1.c). All together, Fig1.a-c illustrates a three-step reversible polar mapping (lossless pixel-to-pixel polar transcoding, stretching and interpolation).

One advantage of using such a polar mapping is that the original pixels within the initial circular iris ring can be traced at any time in the unwrapped versions of the iris.

On the other hand, the extent of the black regions in Fig.1.b is a measure of the difference between an ideal polar mapping (in continuous geometry) and a practical lossless pixel-to-pixel polar mapping defined for digitized images.

The third advantage is the fact that, here, the influence of the pupil dilation on the recognition performance (documented in [19]) is explained and illustrated graphically: comparison of two irides means to overlap two trapezes through an elastic deformation. At least because of the collarette, the deformation in the radial direction is far from uniform. This is the reason for which our Gabor Analytic Iris Texture Binary Encoder [13],[14] parses iris features only in the angular direction.

Fig.2 shows what happens to the vectors A, B and C at steps 4-5 of the CFIS procedure: behind the combined crisp indicator function (crisp membership assignment) of a 3-means quantization (Fig.2), there are fuzzy membership assignment functions defined from the set of lines within the rectangular unwrapped iris area (RUI-Fig.1) to the pupil, to the iris, to the area outside the iris and even to the iris boundaries. The area delimited between the fuzzy iris boundaries is a fuzzy iris band. Its preimage through the polar mapping is a circular fuzzy iris ring.

Three fuzzy iris bands are determined using vectors A, B, C. The final result is computed evaluating the odds that the lines within the unwrapped iris area belong to the actual iris segment. This is done in step 6 of the CFIS procedure by counting the votes received for each line within the unwrapped iris area as a member of a fuzzy iris band.

The most important aspect of the CFIS procedure is that it performs three searches within a 1-dimensional signal whose length equals the radius of the initial iris circular region (Fig.1.a). For example, the dimension of the parameter space which is needed to be searched in order to find the limbic boundary in Fig.2 is: 3*112=336 pixels. On the other hand, using a Hough accumulator with $343=7^3$ cells to extract a circle (limbic boundary) from the edges of an eye image of dimension 240x320 pixels will be totally insufficient, but still computationally more expensive.

IV. THE DEMO PROGRAM

Circular Fuzzy Iris Segmentation demo version [15] is currently implemented in Matlab and can be tested against the entire Bath University Iris Database (free version [20]) which contains 1000 eye images. Basically, the demo program is an implementation of the CFIS procedure with some specific practical adaptations: fault-tolerance, timing, graphical display, etc.

CFIS demo program enables pupil localization in 12 frames per second and limbic boundary localization in 5 frames per second, for eye images of dimension 240x320 pixels. It also leads to the following iris segmentation error rates:
- Total number of failure cases: **6**;
- Pupil finder failures: **1**;
- Limbic boundary detection failures: **5**.

The demo program proves that iris segmentation can be treated as being a 1-dimensional optimization problem if there is enough accurate morphological information stored as chromatic variation.

Another important aspect is that the segmentation results obtained by applying CFIS procedure are confirmed by the recognition results in [13], [14].

## V. Conclusion

This paper introduced combined crisp and fuzzy indicators of a disjoint reunion which are meant to allow a unified dual description of the k-means quantization as a crisp and as a fuzzy entity, respectively.

Both of them are instruments that enable us to view the result of a segmentation as a crisp indicator defined from the input signal to a collection of segments encoded as a list of *arbitrary symbols* (possibly *non-numeric*, and more often found *outside* [0, 1] interval), but also as a fuzzy membership function.

A practical example is the case of Circular Fuzzy Iris Segmentation procedure in which combined crisp and fuzzy indicators are encoded in [1, 3] interval (Fig.2).

ACKNOWLEDGMENT

I wish to thank my PhD Coordinator, *Professor Luminita State* (University of Pitesti, RO) - for the comments, criticism, and constant moral and scientific support during the last two years, and *Professor Donald Monro* (University of Bath, UK) - for granting the access to the Bath University Iris Database.